\documentclass{article}

\usepackage{arxiv}

\usepackage[utf8]{inputenc} % allow utf-8 input
\usepackage[T1]{fontenc}    % use 8-bit T1 fonts
\usepackage{hyperref}       % hyperlinks
\usepackage{url}            % simple URL typesetting
\usepackage{booktabs}       % professional-quality tables
\usepackage{amsfonts}       % blackboard math symbols
\usepackage{nicefrac}       % compact symbols for 1/2, etc.
\usepackage{microtype}      % microtypography
\usepackage{graphicx}
\usepackage{natbib}
\usepackage{doi}

\title{Deep Learning and Synthetic Media}

\date{}

\author{Raphaël Millière}

\author{Raphaël Millière \\
  Center for Science and Society\\
  Columbia University\\
  New York, NY 10027 \\
  \texttt{rm3799@columbia.edu} \\
}

\renewcommand{\headeright}{Forthcoming in \emph{Synthese}}
\renewcommand{\undertitle}{Forthcoming in \emph{Synthese}}
\renewcommand{\shorttitle}{Deep Learning and Synthetic Audiovisual Media}

\hypersetup{
pdftitle={Deep learning and synthetic media},
pdfsubject={cs.LG},
pdfauthor={Raphaël Millière},
pdfkeywords={deepfakes, deep learning, media synthesis, disinformation, art},
}

\begin{document}

\maketitle

% \noindent\fbox{%
%     \parbox{0.7\textwidth}{%
%         This the penultimate draft of the article, shared with the publisher's permission. Please refer to the published version for citation.
%     }%
% }

\begin{abstract}
Deep learning algorithms are rapidly changing the way in which audiovisual media can be produced. Synthetic audiovisual media generated with deep learning -- often subsumed colloquially under the label ``deepfakes'' -- have a number of impressive characteristics; they are increasingly trivial to produce, and can be indistinguishable from real sounds and images recorded with a sensor. Much attention has been dedicated to ethical concerns raised by this technological development. Here, I focus instead on a set of issues related to the notion of synthetic audiovisual media, its place within a broader taxonomy of audiovisual media, and how deep learning techniques differ from more traditional approaches to media synthesis. After reviewing important etiological features of deep learning pipelines for media manipulation and generation, I argue that ``deepfakes'' and related synthetic media produced with such pipelines do not merely offer incremental improvements over previous methods, but challenge traditional taxonomical distinctions, and pave the way for genuinely novel kinds of audiovisual media.
\end{abstract}


\section{Introduction}
\label{intro}

Recent research in artificial intelligence has been dominated by deep learning (DL), a class of algorithms inspired by biological neural networks that can learn automatically to perform certain tasks from large amounts of data. Much work has been dedicated to developing DL algorithms capable of perceiving and exploring (real or virtual) environments, processing and understanding natural language, and even reasoning. Many of these tasks require DL algorithms to learn from audiovisual media, such as audio recordings, images, and videos. However, a significant amount of recent work in DL has also gone towards crafting algorithms that can \emph{synthesize} novel audiovisual media, with remarkable success.

Generating audiovisual media with the help of computers is not new. Software for music, image, and video editing, 3D modeling, or electronic music composition have existed for some time. Computer-generated imagery (CGI) is now ubiquitous in animation and even live-action films, as well as video games. Nonetheless, the recent progress of DL in the domain of audiovisual media synthesis is rapidly changing the way in which we approach media creation, whether for communication, entertainment, or artistic purposes. An impressive and salient example of this progress can be found in so-called ``deepfakes'', a portmanteau word formed from ``deep learning'' and ``fake'' (\cite{tolosanaDeepfakesSurveyFace2020}). This term originated in 2017, from the name of a Reddit user who developed a method based on DL to substitute the face of an actor or actress in pornographic videos with the face of a celebrity. However, since its introduction, the term ``deepfake'' has been generically applied to videos in which faces have been replaced or otherwise digitally altered with the help of DL algorithms, and even more broadly to any DL-based manipulations of sound, image and video. 

While deepfakes have recently garnered attention in philosophy, the discussion has mostly focused on their potentially harmful uses, such as impersonating identities and disseminating false information (\cite{floridiArtificialIntelligenceDeepfakes2018,deruiterDistinctWrongDeepfakes2021}), undermining the epistemic and testimonial value of photographic media and videos (\cite{fallisEpistemicThreatDeepfakes2020,riniDeepfakesEpistemicBackstop2020}), and damaging reputations or furthering gender inequality through fake pornographic media (\cite{ohmanIntroducingPervertDilemma2020}). These ethical and epistemic concerns are significant, and warranted. However, deepfakes and similar techniques also raise broader issues about the notion of synthetic audiovisual media, as DL algorithms appear to challenge traditional distinctions between different kinds of media synthesis.

Given the lack of a single clear definition of deepfakes, it is helpful to focus on the more explicit notion of \emph{DL-based synthetic audiovisual media}, or DLSAM for short. My aim is in this paper is to elucidate how DLSAM fit in the landscape of audiovisual media. To this end, I will propose a broad taxonomy of audiovisual media (\S\ref{sec:taxonomy}), and reflect upon where to place DLSAM alongside traditional media (\S\ref{sec:DLSAM}). Drawing upon the sub-categories of DLSAM informed by this taxonomy, I will discuss the extent to which they represent a qualitative change in media creation that should lead us to expand our understanding of synthetic audiovisual media, or whether they constitute merely incremental progress over traditional approaches (\S\ref{sec:continuity}).

\section{A Taxonomy of Audiovisual Media}
\label{sec:taxonomy}

``Media'' is a polysemous term. It may refer to physical materials (e.g., tape, disk, or paper) used for recording or reproducing data (\emph{storage media}); or, by extension, to the format in which data is stored, such as JPEG or MP3 for digital image and sound respectively (\emph{media format}). It may also refer more broadly to the kinds of data that can be recorded or reproduced in various materials and formats (\emph{media type}); in that sense, text, image, and sound are distinct media, even though they may be stored in the same physical substrate (e.g., a hard drive).\footnote{As noted by an anonymous referee, it may be argued that, in that final sense but not in the others, the information conveyed by a particular medium is partially constituted by that medium.} By ``audiovisual media'', I will generally refer to artifacts or events involving sound, still images, moving images, or a combination of the above, produced to deliver auditory and/or visual information for a variety of purposes, including communication, art, and entertainment. For convenience, I will refer to media involving sound only as ``auditory media'', media involving still images only as ``static visual media'', and media involving moving images (with or without sound) as ``dynamic visual media''.

\begin{figure}
\begin{center}
\includegraphics[width=0.8\textwidth]{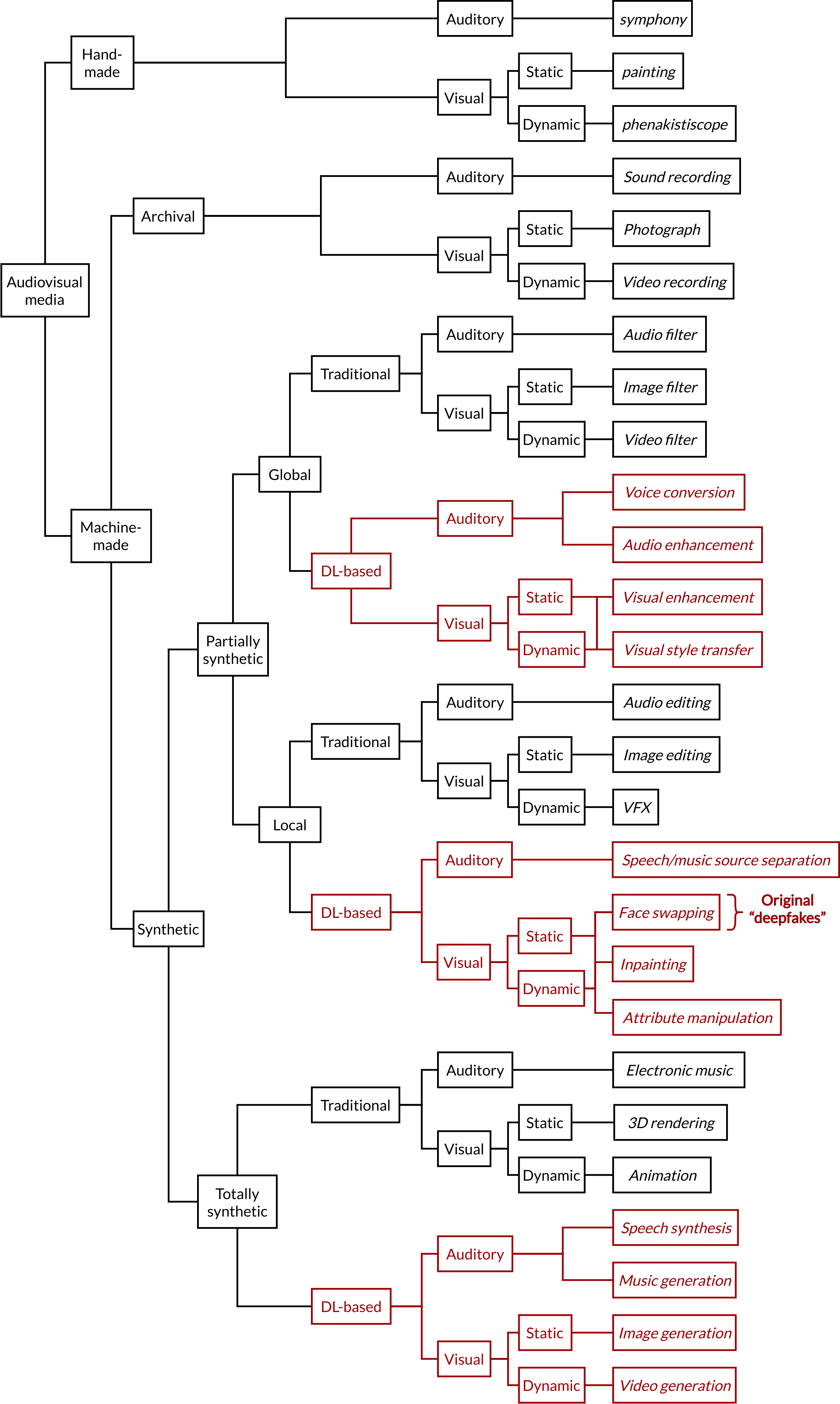}
\caption{A taxonomy of audiovisual media (DLSAM in red).}
\label{fig:taxonomy}
\end{center}
\end{figure}

Audiovisual media fall into two broad etiological categories: \emph{hand-made} media, produced by hand or with the help of manual tools (e.g., paint brushes), and \emph{machine-made} media, produced with the help of more sophisticated devices whose core mechanism is not, or not merely, hand-operated (e.g. cameras and computers) (fig. \ref{fig:taxonomy}). Symphonies, paintings, and phenakistiscopes are examples of hand-made auditory, static visual, and dynamic visual media respectively.

Machine-made audiovisual media can be further divided into two categories: \emph{archival} and \emph{synthetic} media. Archival media are brought about by real objects and events in a mechanical manner. They can be said to ``record'' reality in so far as they capture it through a process that is not directly mediated by the producer's desires and beliefs. More specifically, they are counterfactually dependent upon the real objects or events that bring them about, even if the intentional attitudes of any human involved in producing them are held fixed. Raw (unprocessed) audio recordings, photographs, and video recordings are examples of archival audiovisual media. What someone believes they are capturing when pressing a button on a camera or microphone is irrelevant to what the camera or microphone will in fact record.

This property of archival audiovisual media corresponds more or less to what Kendall Walton has characterized as ``transparency'' with respect to photography in particular (\cite{waltonTransparentPicturesNature1984}). In Walton's terminology, photographs are \emph{transparent} because they put us in perceptual contact with reality: ``we see the world \emph{through} them'' (p. 251). Paintings, no matter how realistic, do not put us in perceptual contact with reality in this way, because they are not mechanically caused by objects in the painter's environment; rather, they might be caused by them only indirectly, through the meditation of the painter's intentional attitudes (e.g., the painter's belief that the object they are attempting to depict looks a certain way).\footnote{There is some debate about whether photographs are actually transparent in Walton's sense (\cite[e.g.,][]{curriePhotographyPaintingPerception1991}).} While Walton focused on the case of archival visual media, it has been argued that raw audio recordings also transparent in that sense (\cite{mizrahiRecordedSoundsAuditory2020}).

By contrast with archival audiovisual media, \emph{synthetic} audiovisual media do not merely record real objects and events; instead, their mode of production intrinsically involves a generative component. In turn, these media can be \emph{partially} or \emph{totally} synthetic (fig. \ref{fig:taxonomy}). The former involve the modification -- through distortion, combination, addition, or subtraction -- of archival media: while they involve a generative component, their also involve source material that has not been generated but recorded. The latter are entirely generative: they involve the creation of new sounds, images, or videos that do not directly incorporate archival media, even if they might be inspired by them.

More specifically, there are two ways in which audiovisual media can be only \emph{partially} synthetic, depending on whether they result from a \emph{global} or \emph{local} manipulation of archival media. Global manipulations involve applying an effect to an entire sound recording, photograph, or video. For example, one can apply a filter to an audio signal to modify its loudness, pitch, frequency ranges, and reverberation, or thoroughly distort it (which is traditionally done with effects pedals and amplifiers in some music genres). In the visual domain, one can also apply a filter to an entire image or video, to adjust various parameters such as hue, brightness, contrast, and saturation, or apply uniform effects like Gaussian blur or noise. By contrast, local manipulations involve modifying, removing, or replacing proper parts of archival audiovisual media instead of adjusting global parameters. For example, one can edit a sound recording to censor parts of it, or add occasional sound effects like a laughing track. In the visual domain, image editing software like Adobe Photoshop can be used to manipulate parts of images, and VFX software can be used to manipulate parts of video recordings through techniques like rotoscoping, compositing, or the integration of computer-generated imagery (CGI).

Totally synthetic audiovisual media are not produced by modifying pre-existing archival media, but consist instead in generating entirely novel sound or imagery. Traditional forms of synthetic media include electronic music and sound effects generated with synthesizers or computers, computer-generated 3D rendering or digital illustration, and animated videos. 

Note that the general distinction between archival and synthetic audiovisual media is orthogonal to the distinction between \emph{analog} and \emph{digital} signals. Analog recording methods store continuous signals directly in or on the media, as a physical texture (e.g., phonograph recording), as a fluctuation in the field strength of a magnetic recording (e.g., tape recording), or through a chemical process that captures a spectrum of color values (e.g., film camera). Digital recording methods involve quantizing an analog signal and representing it as discrete numbers on a machine-readable data storage. Archival media can be produced through either analog (e.g., tape recorder, film camera) or digital (e.g., digital microphone and camera) recording methods. Likewise, some synthetic media can be produced through analog means, as shown by the famous 1860 composite portrait of Abraham Lincoln produced with lithographs of Lincoln's head and of John Calhoun's body.

Some of these taxonomic choices might be debatable. For example, hand-made audiovisual media could be included within the category of synthetic media, as they involve generating new sounds or images. In that case, most hand-made media might be considered as totally synthetic, with some exceptions -- such as painting over a photograph. There are also a few edge cases in which the distinction between archival and synthetic media becomes less obvious, such as artworks involving collages of photographs in which no part of the source material is removed or occluded. By and large, however, the proposed taxonomy is useful as a heuristic to think about different kinds of traditional audiovisual media. 

\section{DL-based Synthetic Audiovisual Media}
\label{sec:DLSAM}

In recent years, the progress of DL in computer science has started transforming the landscape of synthetic audiovisual media. DL is currently the most prominent method in research on artificial intelligence, where it has surpassed more traditional techniques in various domains including computer vision and natural language processing. It is part of a broader family of machine learning methods using so-called \emph{artificial neural networks}, loosely inspired by the mammalian brain, that can learn to represent features of data for various downstream tasks such as detection or classification. Deep learning specifically refers to machine learning methods using \emph{deep} artificial neural networks, whose units are organized in multiple processing layers between input and output, enabling them to efficiently learn representations of data at several levels of abstraction (\cite{lecunDeepLearning2015,bucknerDeepLearningPhilosophical2019}). Deep neural networks can be trained end-to-end: given a large enough training dataset, they can learn automatically how to perform a given task with a high success rate, either through labeled samples (\emph{supervised learning}) or from raw untagged data (\emph{unsupervised learning}).

Given enough training data and computational power, DL methods have proven remarkably effective at classification tasks, such as labeling images using many predetermined classes like ``African elephant'' or ``burrito'' (\cite{krizhevskyImageNetClassificationDeep2017}). However, the recent progress of DL has also expanded to the manipulation and synthesis of sound, image, and video, with so-called ``deepfakes'' (\cite{tolosanaDeepfakesSurveyFace2020}). As mentioned at the outset, I will mostly leave this label aside to focus on the more precise category of \emph{DL-based synthetic audiovisual media}, or DLSAM for short.
Most DLSAM are produced by a subclass of unsupervised DL algorithms called \emph{deep generative models}. Any kind of observed data, such as speech or images, can be thought of as finite set of samples from an underlying probability distribution in a (typically high-dimensional) space. For example, the space of possible color images made of 512x512 pixels has no less than 786,432 dimensions -- three dimensions per pixel, one for each of the three channels of the RGB color space. Any given 512x512 image can be thought of as a point within that high-dimensional space. Thus, all 512x512 images of a given class, such as dog photographs, or real-world photographs in general, have a specific probability distribution within $\mathbb{R}^{786432}$. Any specific 512x512 image can be treated as a sample from an underlying distribution in high-dimensional pixel space. The same applies to other kinds of high-dimensional data, such as sounds or videos.

At a first pass, the goal of deep generative models is to learn the probability distribution of a finite unlabeled dataset on which they are trained. When trained successfully, they can then be used to estimate the likelihood of a given sample, and, importantly, to create \emph{new} samples that are similar to samples from the learned probability distribution (this is the \emph{generative} component of the model). More precisely, deep generative models learn an intractable probability distribution $X$ defined over $\mathbb{R}^{n}$, where $X$ is typically complicated (e.g., disjoint), and $n$ is typically large. A large but finite number of independent samples from $X$ are used as the model's training data. The goal of training is to obtain a generator that maps samples from a \emph{tractable} probability distribution $Z$ in $\mathbb{R}^{q}$ to points in $\mathbb{R}^{n}$ that resemble them, where $q$ is typically smaller than $n$. $Z$ is called the \emph{latent space} of the model. After training, the generator can generate new samples in $X$ (e.g., 512x512 images) from the latent space $Z$.

\begin{figure}
\begin{center}
\includegraphics[width=0.8\textwidth]{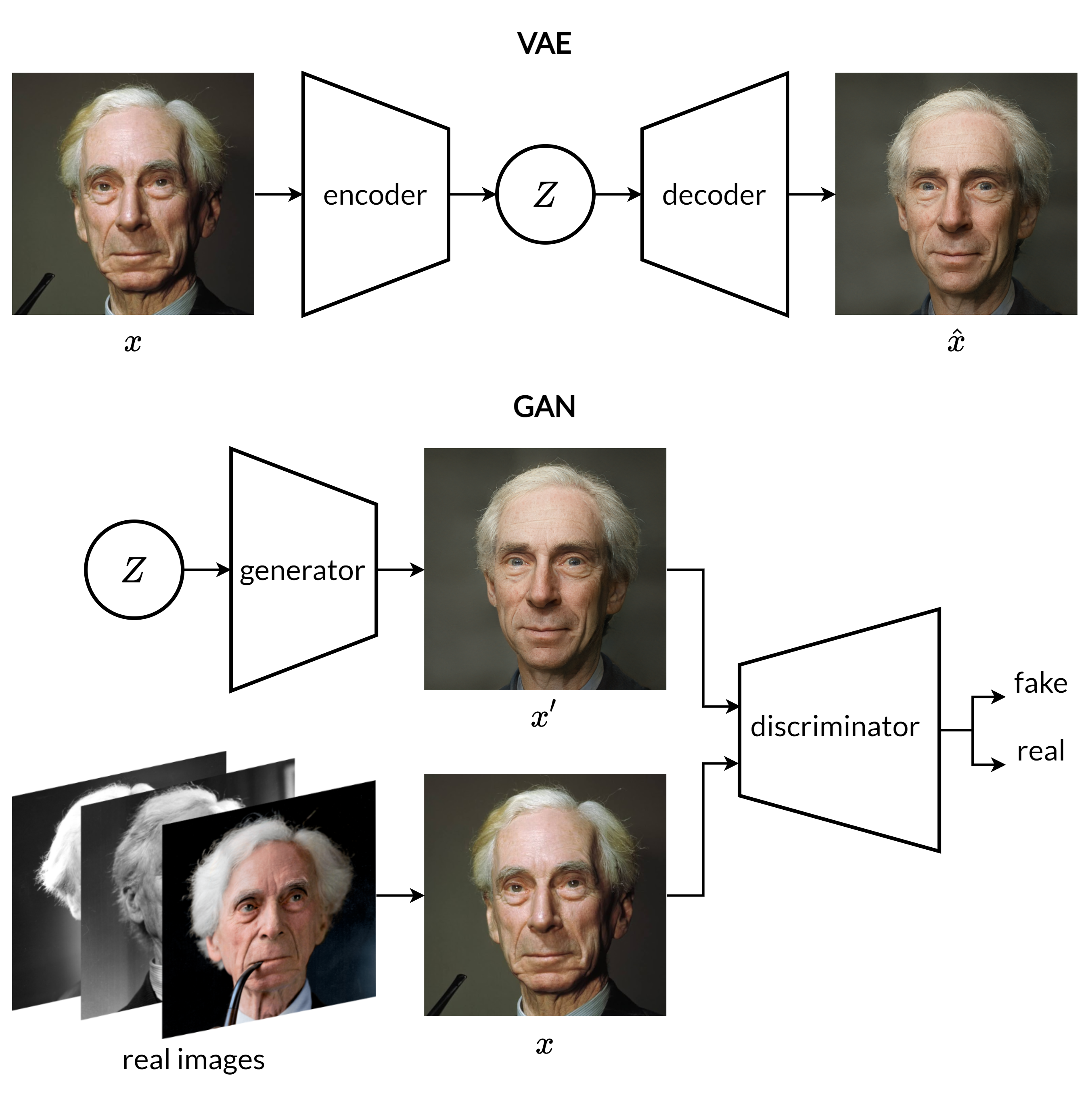}
\caption{The simplified architectures of a VAE and a GAN.}
\label{fig:architectures}
\end{center}
\end{figure}

There are two main types of deep generative models: variational autoencoders (VAEs), used for example in most traditional ``deepfakes'' (\cite{kingmaFastGradientBasedInference2013,pmlr-v32-rezende14}); and generative adversarial networks (GANs), used in other forms of audiovisual media synthesis (\cite{NIPS2014_5ca3e9b1}). VAEs have two parts: an encoder and decoder (fig. \ref{fig:architectures}, top). They learn the probability distribution of the data by encoding training samples into a low-dimensional latent space, then decoding the resulting latent representations to reconstruct them as outputs, while minimizing the difference between real input and reconstructed output. By contrast, GANs have a game theoretic design that includes two different sub-networks, a \textit{generator} and a \textit{discriminator}, competing with each other during training (fig. \ref{fig:architectures}, bottom). The generator is trained to generate new samples, while the discriminator is trained to classify samples as either real (from the training data) or fake (produced by the generator). The generator's objective is to ``fool'' the discriminator into classifying its outputs as real, that is, to increase the error rate of the discriminator. Over time, the discriminator gets better at detecting fakes, and in return samples synthesized by the generator get better at fooling the discriminator. After a sufficient number of training events, the generator can produce realistic outputs that capture the statistical properties of the dataset well enough to look convincing to the discriminator, and often to humans.

The intricacies of deep generative models have significant implications for our understanding the nature of DLSAM, as well as future possibilities for synthetic media. Before discussing these implications, I will give an overview of the main kinds of DLSAM that can be produced with existing DL algorithms. Using the taxonomy introduced in \S\ref{sec:taxonomy}, we can distinguish three categories of DLSAM: (a) global partially synthetic DLSAM, (b) local partially synthetic DLSAM, and (c) totally synthetic DLSAM (fig. \ref{fig:taxonomy}). As we shall see, the original ``deepfakes'', consisting in replacing faces in videos, can be viewed as instances of the second category of DLSAM, and hardly span the full spectrum of DL-based methods to alter or generate audiovisual media. While it is useful, at a first approximation, to locate different types of DLSAM within the traditional taxonomy of audiovisual media, it will become apparent later that a deeper understanding of their etiology challenges some categorical distinctions upon which this taxonomy is premised.

\subsection{Global partially synthetic DLSAM}

In this first category are audiovisual media produced by altering global properties of existing media with the help of DL algorithms. Examples in the auditory domain include audio enhancement and voice conversion. Audio enhancement straightforwardly consists in enhancing the perceived quality of an audio file, which is especially useful for noisy speech recordings (\cite{huDCCRNDeepComplex2020}), while voice conversion consists in modifying the voice of the speaker in a recording to make them sound like that of another speaker without altering speech content (\cite{huangSequencetoSequenceBaselineVoice2020}).

\begin{figure}
\begin{center}
\includegraphics[width=0.7\textwidth]{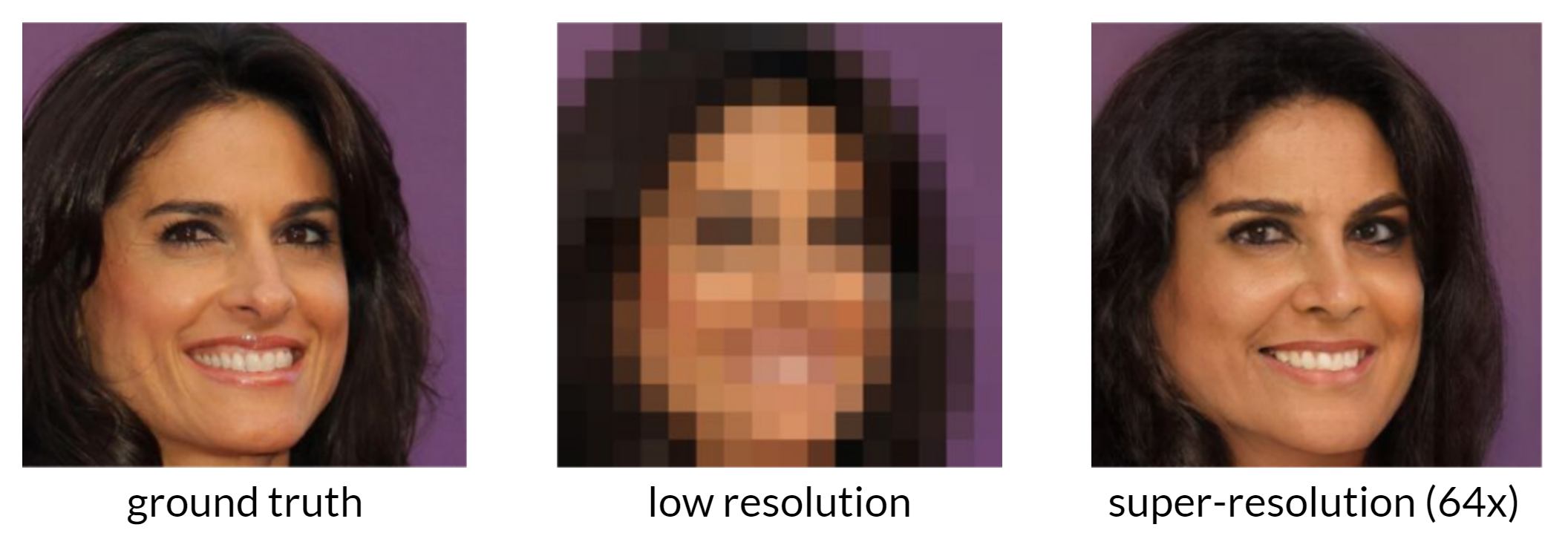}
\caption{An example of super-resolution (adapted from \cite{chanGLEANGenerativeLatent2021}).}
\label{fig:SR}
\end{center}
\end{figure}

In the visual domain, this category includes both static and dynamic visual enhancement and style transfer. Like its auditory counterpart, visual enhancement consists in improving the perceived quality of images and videos. It encompasses ``denoising'', or removing noise from images/videos (\cite{zhangGaussianDenoiserResidual2016}); reconstituting bright images/videos from sensor data in very dark environments (\cite{chenLearningSeeDark2018}); ``debluring'', or removing visual blur (\cite{kupynDeblurGANBlindMotion2018}); restoring severely degraded images/videos (\cite{wanOldPhotoRestoration2020}); ``colorization'', or adding colors to black-and-white images/videos (\cite{kumarColorizationTransformer2020}); and ``super-resolution'', or increasing the resolution of images/videos to add missing detail (\cite{ledigPhotoRealisticSingleImage2017}, see fig. \ref{fig:SR}). Visual style transfer consists in changing the style of an image/video in one domain, such as a photograph, to the style of an image/video in another domain, such as a painting, while roughly preserving its compositional structure (\cite{gatysNeuralAlgorithmArtistic2015}).

\subsection{Local partially synthetic DLSAM}

Audiovisual media produced by altering local properties of existing media with DL algorithms can be subsumed under this second category of DLSAM. In the auditory domain, this concerns in particular audio files produced through source separation. Speech source separation consists in extracting overlapping speech sources in a given mixed speech signal as separate signals (\cite{subakanAttentionAllYou2021}), while music source separation consists in decomposing musical recordings into their constitutive components, such as generating separate tracks for the vocals, bass, and drums of a song (\cite{spleeter2020}). Auditory media produced through source separation are instances of \emph{local} partially synthetic media, insofar as they involve removing parts of a recording while preserving others, instead of applying a global transformation to the recording as a whole.

\begin{figure}
\begin{center}
\includegraphics[width=0.7\textwidth]{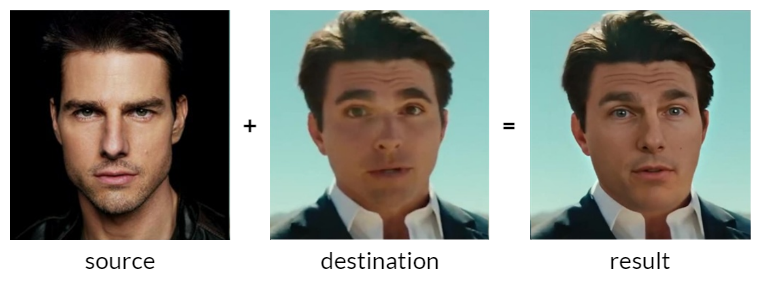}
\caption{Face swapping ``deepfake'' (adapted from \cite{perovDeepFaceLabIntegratedFlexible2021}).}
\label{fig:deepfakes}
\end{center}
\end{figure}

In the visual domain, this category encompasses images and videos produced through ``deepfakes'' in the narrow sense, as well as inpainting, and attribute manipulation. In this context, ``deepfake'' refers to face swapping, head puppetry, or lip syncing. Face swapping is the method behind the original meaning of the term (\cite{tolosanaDeepfakesSurveyFace2020}). It consists in replacing a subject's face in images or videos with someone else's (fig. \ref{fig:deepfakes}). State-of-the-art pipelines for face swapping are fairly complex, involving three steps: an extraction step to retrieve faces from sources images and from the target image or video (this requires a mixture of face detection, facial landmark extraction to align detected faces, and face segmentation to crop them from images); a training step that uses an autoencoder architecture to create latent representations of the source and target faces with a shared encoder; and a conversion step to re-align the decoded (generated) faces with the target, blend it, and sharpen it (\cite{perovDeepFaceLabIntegratedFlexible2021}).

Head puppetry or ``talking head generation'' is the task of generating a plausible video of a talking head from a source image or video by mimicking the movements and facial expressions of a reference video (\cite{zakharovFewShotAdversarialLearning2019}), while lip syncing consists in synchronizing lip movements on a video to match a target speech segment (\cite{prajwalLipSyncExpert2020}). Head puppetry and lip syncing are both forms of \emph{motion transfer}, which refers more broadly to the task of mapping the motion of a given individual in source video to the motion of another individual in a target image or video (\cite{zhuProgressiveAlignedPose2021,kappelHighFidelityNeuralHuman2021}). Face swapping, head puppetry, and lip syncing are commonly referred to as ``deepfakes'' because they can be used to usurp someone's identity in a video; however, they involve distinct generation pipelines.

\begin{figure}
\begin{center}
\includegraphics[width=\textwidth]{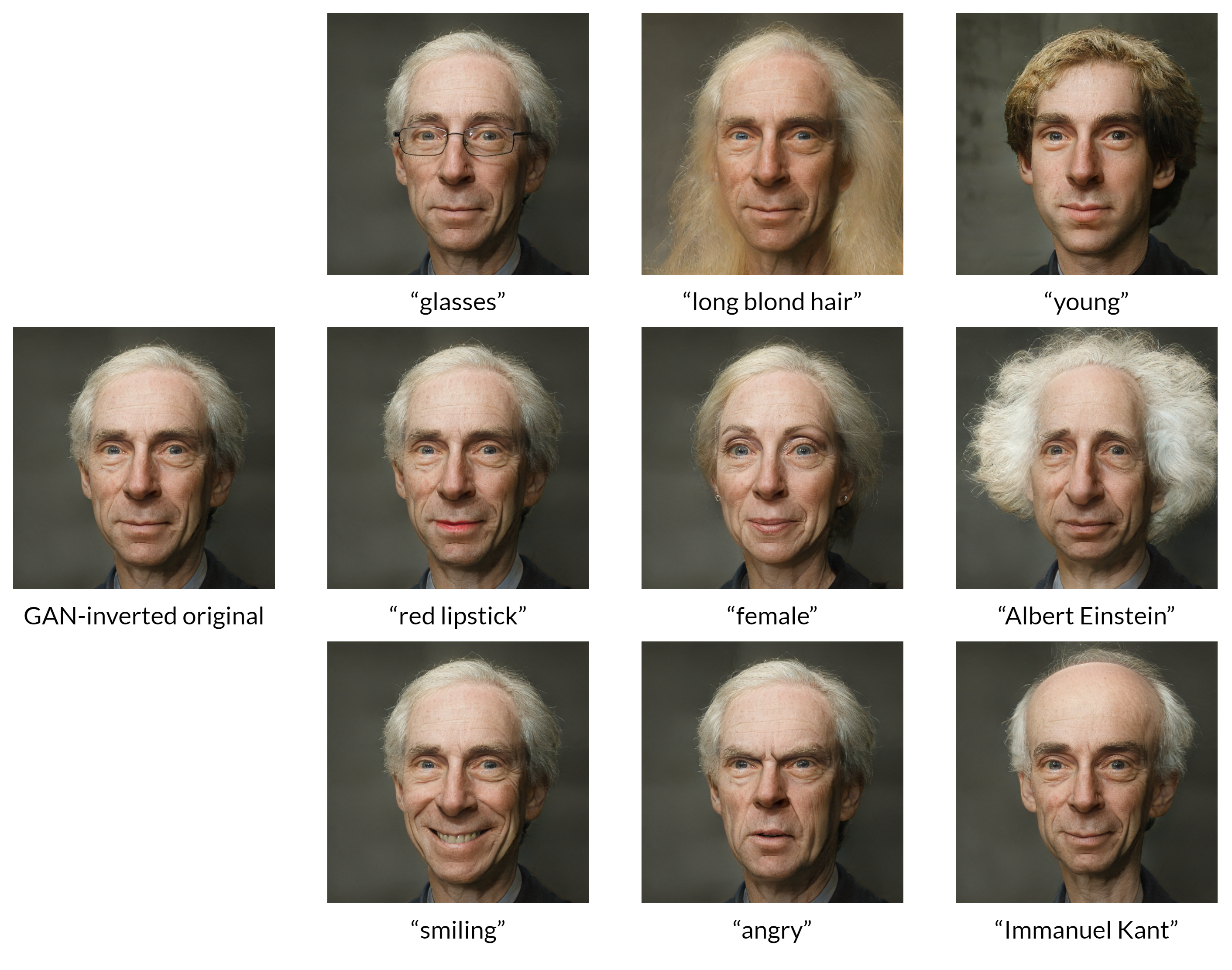}
\caption{Semantic face editing of a photograph of Bertrand Russell with text prompts, produced with StyleCLIP (\cite{patashnikStyleCLIPTextDrivenManipulation2021}).}
\label{fig:russell}
\end{center}
\end{figure}

Inpainting involves reconstructing missing regions in an image or video sequence with contents that are spatially and temporally coherent (\cite{yuFreeFormImageInpainting2019,xuDeepFlowGuidedVideo2019}). Finally, attribute manipulation broadly refers to a broad range of techniques designed to manipulate local features of images and videos. Semantic face editing or facial manipulation consists in manipulating various attributes in an headshot, including gender, age, race, pose, expression, presence of accessories (eyewear, headgear, jewelry), hairstyle, hair/skin/eye color, makeup, as well as the size and shape of any part of the face (ears, nose, eyes, mouth, etc.) (\cite{leeMaskGANDiverseInteractive2020,shenInterpretingLatentSpace2020,viazovetskyiStyleGAN2DistillationFeedforward2020}). Similar techniques can be used to manipulate the orientation, size, color, texture, and shape of objects in an image more generally (\cite{shenClosedFormFactorizationLatent2021}). As we shall see, this can even be done by using a linguistic description to guide the modification of high-level and abstract properties of persons or objects in an image, e.g. adding glasses to a photograph of a face with the caption ``glasses'' (\cite{patashnikStyleCLIPTextDrivenManipulation2021}, fig. \ref{fig:russell}).

\subsection{Totally synthetic DLSAM}

In this last category are audiovisual media entirely synthesized with the help of DL algorithms, rather than produced by altering pre-existing media. Such synthesis can be \emph{conditional}, when samples are generated conditionally on labels from the dataset used for training, or \emph{unconditional}, when samples are generated unconditionally from the dataset. In the auditory domain, totally synthetic DLSAM include speech synthesis, which consists in generating speech from some other modality like text (text-to-speech) or lip movements that can be conditioned on the voice of a specific speaker (\cite{shenNaturalTTSSynthesis2018}); and music generation, which consists in generating a musical piece that can be conditioned on specific lyrics, musical style or instrumentation (\cite{dhariwalJukeboxGenerativeModel2020}).

In the visual domain, this category includes image and video generation. These can also be unconditional (\cite{vahdatScorebasedGenerativeModeling2021,tianGoodImageGenerator2021}), or conditioned for example on a specific class of objects (e.g., dogs) from a dataset (\cite{brockLargeScaleGAN2019}), on a layout describing the location of the objects to be included in the output image/video (\cite{sylvainObjectCentricImageGeneration2020}), or on a text caption describing the output image/video (\cite{rameshZeroShotTexttoImageGeneration2021}).

The progress of image generation has been remarkable since the introduction of GANs in 2014. State-of-the-art GANs trained on domain-specific datasets, such as human faces, can now generate high-resolution photorealistic images of non-existent people, scenes, and objects.\footnote{See \url{https://www.thispersondoesnotexist.com} for random samples of non-existent faces generated with StyleGAN2 (\cite{karrasAnalyzingImprovingImage2020}), and \url{https://thisxdoesnotexist.com} for more examples in other domains. By "photorealistic images", I mean images that the average viewer cannot reliably distinguish from genuine photographs.} The resulting images are increasingly difficult to discriminate from real photographs, even for human faces, on which we are well-attuned to detecting anomalies.\footnote{The reader can try guessing which of two images was generated by a GAN on \url{https://www.whichfaceisreal.com} (note that the model used for generating these images is no longer the state of the art for image generation).} Other methods now achieve equally impressive results for more varied classes or higher-resolution outputs, such as diffusion models (\cite{dhariwalDiffusionModelsBeat2021}) and Transformer models (\cite{esserTamingTransformersHighResolution2021}).

It has also become increasingly easy to guide image generation directly with text. DALL-E, a new multimodal Transformer model trained on a dataset of text–image pairs, is capable of generating plausible images in a variety of styles simply from a text description of the desired output (\cite{rameshZeroShotTexttoImageGeneration2021}). DALL-E's outputs can exhibit complex compositional structure corresponding to that of the input text sequences, such as ``An armchair in the shape of an avocado'', ``a small red block sitting on a large green block'', or ``an emoji of a baby penguin wearing a blue hat, red gloves, green shirt, and yellow pants''. DALL-E has been developed jointly with another multimodal model, called CLIP, capable of producing a natural language caption for any input image (\cite{radfordLearningTransferableVisual2021}). Using CLIP to steer the generation process, it is also possible to produce images with GANs from natural language descriptions of the desired output (\cite{galatoloGeneratingImagesCaption2021,patashnikStyleCLIPTextDrivenManipulation2021}; see fig. \ref{fig:russell}).

If these trends continue -- and there is no reason for them to slow down significantly as hardware improvement and architectural breakthroughs continue to spur larger and more efficient models --, it is only a matter of time before DL algorithms allow us to generate high-resolution stylized or photorealistic samples of arbitrary scenes that are consistently indistinguishable from human-made outputs. In the domain of static visual media, that goal is already within sight for medium to large resolutions (around 1024x1024 pixels at the time of writing). The domain of dynamic visual media presents a significantly greater challenge, as the spatiotemporal consistency of the scene needs to be taken into account. Nonetheless, it is plausible that we will be able to synthesize realistic and coherent video scenes at relatively high resolution in the short to medium term, beyond mere face swapping in existing videos. 

\section{The Continuity Question}
\label{sec:continuity}

DL has expanded the limits of synthetic audiovisual media beyond what was possible with previously available methods. Nonetheless, the extent to which DLSAM really differ from traditional synthetic audiovisual media is not immediately clear, aside from the obvious fact that they are produced with deep artificial neural networks. After all, each of the three categories of DLSAM distinguished in the previous section -- \emph{global partially synthetic}, \emph{local partially synthetic}, and \emph{totally synthetic} -- has traditional counterparts that do not involve DL. In fact, it is not implausible that many DLSAM, from visual enhancement to face swapping, could be copied rather closely with more traditional methods, given enough time, skills and resources. DL undoubtedly cuts down the resource requirements for the production of synthetic media, making them easier and faster to generate. But one might wonder whether the difference between DLSAM and traditional media is merely one of performance and convenience, lowering the barrier to entry for the media creation; or whether there are additional differences that warrant giving a special status to DLSAM in the landscape of audiovisual media. More specifically, the question is whether DLSAM simply make it easier, faster, and/or cheaper to produce audiovisual media that may have otherwise been produced through more traditional means; or whether they also enable the production of new forms of audiovisual media that challenge traditional categories. Let us call this the Continuity Question. 

There are two ways to understand the claim that DLSAM might be discontinuous with -- or qualitatively different from -- traditional audiovisual media. On an epistemic reading, DLSAM may not threaten the taxonomy itself, but simply make it more difficult for media consumers to tell where a specific instance of DLSAM lies within that taxonomy. On an ontological reading, DLSAM genuinely challenges the taxonomy itself, by undermining some categorical distinctions on which it is premised. The epistemic reading of the discontinuity claim is clearly correct if one insists on making all instances of DLSAM fit within the traditional taxonomy of audiovisual media. For example, it is increasingly easy to mistake photorealistic GAN-generated images of human faces (in the category of \emph{totally synthetic media}) for actual photographs of human faces (in the category of \emph{archival media}). It should be noted, however, that traditional methods of media manipulation also enable such confusions; DL-based techniques merely make it easier to fool media consumers into misjudging the source of an item. In that respect, DLSAM do not represent such a radical departure from previous approaches to media creation. 

In what follows, I will focus instead on the ontological reading of the Continuity Question. I will review a number of ways in which DLSAM differ from other audiovisual media, and ask whether any of these differences genuinely threatens the traditional taxonomy. Answering this question largely depends on the choice of criteria one deems relevant to distinguishing kinds of audiovisual media. So far, I have mostly considered etiological criteria, which form the basis of the taxonomy illustrated in fig. \ref{fig:taxonomy}. If we leave etiology aside, and simply consider the auditory and pictorial properties of DLSAM, it seems difficult to see how they really differ from traditional media. It is also doubtful that DLSAM can be squarely distinguished from traditional media with respect to their intended role, be it communication, deception, art, or entertainment -- all of which are also fulfilled by traditional methods. However, I will argue that the etiology of DLSAM does play a crucial role in setting them apart from other kinds of audiovisual media, for a few different reasons, beyond the surface-level observation that they are produced with DL algorithms. Once we have a better understanding of how DSLAM are produced, it will become clear that they challenge some categorical distinctions of the traditional taxonomy. 

\subsection{Resource requirements}

One respect in which DLSAM clearly differ from traditional synthetic media is the resources they require for production. First, deep learning techniques have considerably lowered the bar for the level of technical and artistic skills required to manipulate or synthesize audiovisual media. Virtually no artistic skills are required for to make a photograph look like a line drawing or a Van Gogh painting with style transfer, to produce a rock song in the style and voice of Elvis Presley using a variational autoencoder,\footnote{See \url{https://openai.com/blog/jukebox} for an example.} to change the eye color, hairstyle, age, or gender of a person in a photograph,\footnote{See \url{https://www.nytimes.com/interactive/2020/11/21/science/artificial-intelligence-fake-people-faces.html}.} or to generate an abstract artwork with a GAN.\footnote{See \url{https://thisartworkdoesnotexist.com}. Of course, curating aesthetically interesting images from thousands of generated samples still requires a trained eye, just like the easy of use of digital photography does not make everyone an artist.} 

For a while, advanced technical skills and access to powerful and expensive hardware were still required to produce DLSAM. When the original ``deepfakes'' came onto the scene in 2017, for example, it was far from trivial to generate them without prior expertise in programming and deep learning. However, this changed dramatically over the past few years. Many smartphone apps now leverage deep learning to manipulate audiovisual media with the click of a button. Popular social media apps like Instagram, Snapchat, and Tiktok offer a broad range of DL-based filters that automatically apply complex transformations to images and videos, including so-called ``beauty filters'' made to enhance the appearance of users. Likewise, third-party standalone apps such as FaceApp and Facetune are entirely dedicated to the manipulation of image and video ``selfies'', using DL to modify specific physical features of users ranging from age and gender to the shape, texture, and tone of body parts. Using these apps requires no special competence beyond basic computer literacy. 

More polished and professional results can also be achieved with user-friendly computer software. For example, DeepFaceLab is a new software designed by prominent deepfake creators that provides ``an easy-to-use way to conduct high-quality face-swapping'' (\cite{perovDeepFaceLabIntegratedFlexible2021}). DeepFaceLab has an accessible user interface and allows anyone to generate high-resolution deepfakes without any coding skills. NVIDIA recently released Canvas, a GAN-powered software to synthesize photorealistic images given a simple semantic layout painted in very broad strokes by the user (\cite{parkSemanticImageSynthesis2019}).\footnote{The method can also be tried in this online demo: http://nvidia-research-mingyuliu.com/gaugan.} Even Photoshop, the most popular software for traditional image manipulation, now includes ``neural filters'' powered by deep learning (\cite{clarkPhotoshopNowWorld2020}). These pieces of software are not merely used for recreational purposes, but also by creative professionals who can benefit from the efficiency and convenience of DL-based techniques. 

The recent progress of text-to-image generation discussed in the previous section is a further step towards making sophisticated audiovisual media manipulation and synthesis completely trivial. Instead of requiring users to fiddle with multiple parameters to modify or generate images and videos, it allows them to merely describe the desired output in natural language. As we have seen, GAN-based algorithms like StyleCLIP enable users to change various attributes of a subject in a photograph with simple captions (fig. \ref{fig:russell}, \cite{patashnikStyleCLIPTextDrivenManipulation2021}), and can even be used through an easy user interface without programming skills.\footnote{See https://youtu.be/5icI0NgALnQ for some examples.} Other previously mentioned methods based on multimodal Transformer models allow the synthesis of entirely novel images from text input (\cite{rameshZeroShotTexttoImageGeneration2021}). A similar procedure has been successfully applied to the manipulation of videos through simple text prompts (\cite{skorokhodovStyleGANVContinuousVideo2021}). It is plausible that further progress in this area will eventually enable anyone to modify or generate any kind of audiovisual media in very fine detail, simply from natural language descriptions.

While deep learning techniques often come at a high computational cost, many kinds of DLSAM can now be created with consumer hardware. For example, the aforementioned smartphone apps offload much of the DL processing to the parent companies' servers, thereby considerably reducing the computational requirements on the users' devices. Almost any modern smartphone can run these apps, and produce close to state-of-the-art DLSAM in a variety of domains, at no additional computational cost. Given these developments, it is no surprise that consumer-facing companies like Facebook, Snap, ByteDance, Lightricks, Adobe, and NVIDIA -- the developers of Instagram, Snapchat, Tiktok, Facetune, Photoshop, and Canvas respectively -- are at the forefront of fundamental and applied research on DL-based computer vision and image/video synthesis (e.g., \cite[][]{heMaskRCNN2017,tianGoodImageGenerator2021,yuFreeFormImageInpainting2019,halperinEndlessLoopsDetecting2021,patashnikStyleCLIPTextDrivenManipulation2021,karrasAliasFreeGenerativeAdversarial2021}).

Taken together, the lowered requirements on artistic skill, technical competence, computational power, and production time afforded by deep learning algorithms have dramatically changed the landscape of synthetic audiovisual media. Using traditional audio, image, and video editing software to obtain comparable results would require, in most cases, considerably more time and skill. What used to require expertise and hard labor can now be done in a few clicks with consumer-level hardware. By itself, this impressive gap does not seem to challenge traditional taxonomical distinctions: DL algorithms make synthetic media easier and faster to produce, but this does not entail their output do not fit squarely within existing media categories. One might be tempted to compare this evolution to the advent of digital image editing software in the 2000s, which made retouching photographs easier, faster, and more effective than previous analog techniques based on painting over negatives. However, this comparison would be selling the novelty of DLSAM short. The range of generative possibilities opened up by deep learning in the realm of synthetic media far outweighs the impact of traditional editing software.

\subsection{New generative possibilities}

I suggested earlier that some DLSAM could be imitated to some degree with traditional techniques given enough time and means, including professional tools and expertise. While this has been true for a long time, the gap between what can be achieved with and without DL algorithms is widening rapidly, and an increasing number of DLSAM simply se5em impossible to produce with traditional methods, no matter the resources available. 

A good example of this evolution is the manipulation of facial features in dynamic visual media, which was the purpose of the original deepfakes. The progress of VFX has led the film industry to experiment with this kind of manipulation over the past few years, to recreate the faces of actors and actresses who cannot be cast in a movie, or digitally rejuvenate cast members. Thus, \emph{Rogue One} (2016) features scenes with Peter Cushing, or rather a posthumous computer-generated duplicate of Cushing painstakingly recreated from individual frames -- the actor having passed away in 1994. Likewise, Denis Villeneuve's \emph{Blade Runner 2049} (2017) includes a scene in which the likeness of Sean Young's character in the original \emph{Blade Runner} (1982) is digitally added. For \emph{The Irishman} (2019), Martin Scorsese and Netflix worked extensively with VFX company Industrial Light \& Magic (ILM) to digitally ``de-age'' Robert De Niro, Al Pacino, and Joe Pesci for many scenes of the movie. The VFX team spent a considerable amount of time studying older movies featuring these actors to see how they should look at various ages. They shot the relevant scenes with a three-camera rig, and used a special software to detect subtle differences in light and shadows on the actors' skin as reference points to carefully replace their faces with younger-looking computer-generated versions frame by frame.

These examples of digital manipulation are extraordinarily costly and time-consuming, but certainly hold their own against early DL-based approaches rendered at low resolutions. State-of-the-art deepfakes, however, have become very competitive with cutting-edge VFX used in the industry. In fact, several deepfakes creators have claimed to produce \emph{better} results at home  in just a few hours of work on consumer-level hardware than entire VFX teams of blockbuster movies with a virtually unlimited budget and months of hard work.\footnote{https://www.forbes.com/sites/petersuciu/2020/12/11/deepfake-star-wars-videos-portent-ways-the-technology-could-be-employed-for-good-and-bad} These results are so impressive, in fact, that one of the most prominent deepfakes creator, who goes by the name ``Shamook'' on Youtube, was hired by ILM in 2021. The paper introducing DeepFaceLab is also explicitly targeted at VFX professional in addition to casual creators, praising the software's ability to ``achieve cinema-quality results with high fidelity'' and emphasizing its potential ``high economic value in the post-production industry [for] replacing the stunt actor with pop stars'' (\cite{perovDeepFaceLabIntegratedFlexible2021}).

Other forms of partial audiovisual synthesis afforded by DL algorithms are leaving traditional techniques behind. Text-to-speech synthesis sounds much more natural and expressive with DL-based approaches, and can be used to convincingly clone someone's voice (\cite{shenNaturalTTSSynthesis2018}). These methods, sometimes referred to as ``audio deepfakes'', have been recently used among real archival recordings in a 2021 documentary about Anthony Bourdain, to make him posthumously read aloud an email sent to a friend. The fake audio is not presented as such, and is seamless enough that it was not detected by critics until the director confessed to the trick in an interview (\cite{rosnerEthicsDeepfakeAnthony2021}). Voice cloning can also be combined with face swapping or motion transfer to produce convincing fake audiovisual media of a subject saying anything (\cite{thiesNeuralVoicePuppetry2020}). Traditional methods simply cannot achieve such results at the same level of quality, let alone in real time.

Total synthesis is another area where DL has overtaken other approaches by a wide margin. It is extremely difficult to produce a completely photorealistic portrait of an arbitrary human face or object in high-resolution from scratch without deep generative models like GANs. Dynamic visual synthesis is even more of a challenge, and even professional 3D animators still struggle to achieve results that could be mistaken for actual video footage, especially when humans are involved (hence the stylized aesthetic of most animated movies). The progress of DL opens up heretofore unimaginable creative possibilities. This remark goes beyond DL's ability to generate synthetic audiovisual media of the same kind as those that produced with traditional methods, simply with incremental improvements in quality -- e.g. resolution, detail, photorealism, or spatiotemporal consistency. DL challenges our taxonomy of audiovisual media, and paves the way for new kinds of synthetic media that defy conventional boundaries.

\subsection{Blurred lines}

I have previously discussed how DLSAM would fit in a taxonomy of audiovisual media that encompasses all traditional approaches (see fig. \ref{fig:taxonomy}). While this is helpful to distinguish different categories of DLSAM at a high level of abstraction, instead of conflating them under the generic label ``deepfake'', the boundaries between these categories can be challenged under closer scrutiny. Indeed, DL-based approaches to media synthesis have started blurring the line between partially and totally synthetic audiovisual media in novel and interesting ways. 

First, DLSAM arguably straddle the line between partial and total synthesis insofar as they are never generated \emph{ex vacuo}, but inherit properties of the data on which the models that produced them were trained. In order to produce convincing outputs in a given domain (e.g., photorealistic images of human faces), artificial neural networks must be trained on a vast amount of samples from that domain (e.g., actual photographs of human faces). Given this training procedure, there is a sense in which even DLSAM generated unconditionally by DL models such as GANs are not quite \emph{totally} synthetic, insofar as they leverage properties of preexisting images present in their training data, and seamlessly recombine them in coherent ways. Depending on how closely a given output reproduces features from the model's training data, this process can resemble partial synthesis. In fact, it is possible for a GAN to produce an output almost identical to one of its training samples. This is likely to happen if the model suffers from ``overfitting'', namely if it simply memorizes samples from its dataset during training, such that the trained model outputs a near copy of one of the training samples, rather than generating a genuinely novel sample that looks like it came from the same probability distribution. Since there is no robust method to completely rule out overfitting with generative models, one cannot determine \emph{a priori} how different a synthesized human face, for example, will really be from a preexisting photograph of a human face present in the training data.

More generally, DL-based audiovisual synthesis can vary widely in how strongly it is conditioned on various parameters, including samples of synthesized or real media. For example, video generation is often conditioned on a preexisting image (\cite{liuInfiniteNaturePerpetual2020}). This image can be a real photograph, in which case it seems closer to a form of partial synthesis; but it can also be a DL-generated image (e.g., the output of a GAN). Thus, the link between the outputs of deep generative models and preexisting media can be more or less distant depending on the presence of conditioning, its nature, and the overall complexity of the generation pipeline.

The distinction between partially and totally synthetic media is also challenged by DLSAM in the other direction, to an even greater extent, when considering examples that I have previously characterized as instances of partially synthetic media. Many examples of state-of-the-art DL-based visual manipulation rely on reconstructing an image with a deep generative model, in order to modify some of its features by adjusting the model's encoding of the reconstructed image before generating a new version of it. The initial step is called ``inversion'': it consists in projecting a real image into the latent embedding space of a generative model (\cite{xiaGANInversionSurvey2021,abdalImage2StyleGANHowEmbed2019,richardsonEncodingStyleStyleGAN2021}). Every image that can be generated by a generative model corresponds to a vector in the model's latent space. In broad terms, inverting a real image $i$ into the latent space $Z$ of a generative model consists finding the vector $z$ that matches $i$ most closely in $Z$, by minimizing the difference between $i$ and the image generated from $z$ (see fig. \ref{fig:architectures}).

\begin{figure}
\begin{center}
\includegraphics[width=0.5\textwidth]{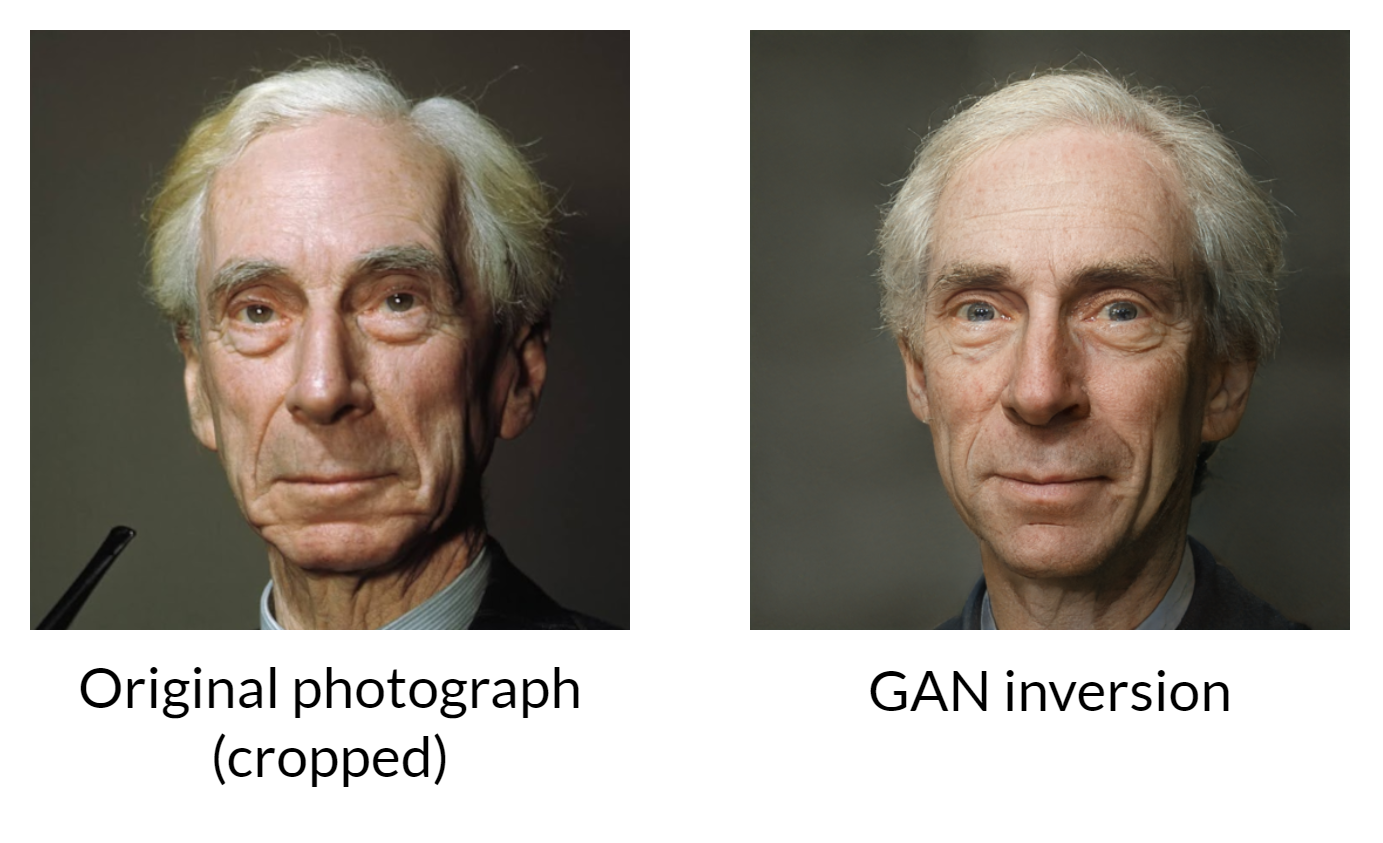}
\caption{A photograph of Russell and its inversion in the latent space of StyleGAN.}
\label{fig:inversion}
\end{center}
\end{figure}

In fig. \ref{fig:russell}, for example, the ``original'' image on the left is not actually a photograph of Bertrand Russell, but the reconstruction of such a photograph produced by inverting it into the latent space of a GAN trained on human faces. The original photograph and its GAN inversion are presented side-by-side in fig. \ref{fig:inversion}. The two images clearly differ in various respects: the pipe that Russell holds in his hand in the original photograph is removed, and various details of the lighting, hair, eyes, nose, ears, lower face, and neck are slightly altered. While the likeness of Russell is rather well captured, the imitation is noticeably imperfect. One could not accurately describe the image on the right as a photograph of Bertrand Russell.

Characterizing images produced by manipulating features through GAN inversion as only \emph{partially} synthetic is somewhat misleading, to the extent that the images were entirely generated by the model, and do not embed any actual part of the real photograph that inspired them. There is a significant difference between this process and the form of local partial synthesis at play, for example, in similar manipulations using traditional software like Photoshop. In the latter case, one genuinely starts from a real photograph to modify or overlay some features (e.g., add some glasses); the final results include a significant proportion of the original image, down to the level of individual pixels (unless the compression or resolution was changed). With GAN-based image manipulation, by contrast, what is really modified is not the original image, but its ``inverted'' counterpart which merely \emph{resembles} it. The resemblance can be near perfect, but there are often noticeable differences between an image and its GAN inversion, as in fig. \ref{fig:inversion}.

DL algorithms are also blurring the line between archival and synthetic media. Many smartphones now include an automatic DL-based post-processing pipeline that artificially enhances images and videos captured with the camera sensor, to reduce noise, brighten the image in low-light condition, fake a shallow depth of field by emulating background blur (so-called ``portrait mode''), and/or increase the media's resolution. Audiovisual media captured with these devices hardly qualifies as ``archival'' in the traditional sense. While there is always a layer of post-processing reflecting technical and aesthetic preferences in going from raw sensor data to an audio recording, image, or video, these DL-based pipeline go further in augmenting the sensor data with nonexistent details synthesized by generative models. Super-resolution is a good example of that process (fig. \ref{fig:SR}). In principle, the goal of super-resolution is simply to increase the resolution of an image or video. In practice, however, even when there is a ground truth about what an image/video would look like at higher resolution (i.e., if it has been downscaled), it is virtually impossible for a super-resolution algorithm to generate a pixel-perfect copy of the higher-resolution version. While the image reconstructed from the very pixelated input in fig. \ref{fig:SR} is undoubtedly impressive, and strikingly similar to the ground truth, it is far from identical to it. If all images produced by a camera were going through such a super-resolution pipeline, it would be misleading to characterize them as archival visual media in the sense defined earlier. In the age of ``AI-enhanced'' audiovisual media, archival and synthetic media increasingly appear to fit on a continuum rather than in discrete categories.

This remark is reminiscent of Walton's suggestion that depictions created through mechanical means may exhibit various \emph{degrees} of transparency depending on their production pipeline (\cite{waltonTransparentPicturesNature1984}). On Walton's view, photographs created by combining two negatives are only partially transparent, because they don't have the right kind of mechanical contact with the constructed scene, but they do with the scenes originally depicted by each negative. Likewise, he argues that an overexposed photograph displays a lower degree of transparency than a well-exposed one, or a grayscale photograph than one in color, etc. These are still instances of ``seeing through'' the medium, as opposed to hand-made media. As Walton puts it:

\begin{quote}
Most photographic constructions are transparent in some of their parts or in certain respects. [...] To perceive things is to be in \emph{contact} with them in a certain way. A mechanical connection with something, like that of photography, counts as contact, whereas a humanly mediated one, like that of painting, does not. (\cite[][pp. 43-4]{waltonTransparentPicturesNature1984})
\end{quote}

On this view, one can see a scene through a photograph -- be in ``contact'' with it -- to various degrees, but most photographic manipulations do not make the output entirely opaque. What applies to analog photographic manipulation should also apply, \emph{mutatis mutandis}, to digital manipulation with traditional editing software. But how do DLSAM fit in that view? They seem to put further pressure on a sharp divide between transparency and opacity. A photograph taken by a smartphone, and automatically processed through DL-based denoising and super-resolution algorithms, does appear to put us in contact with reality \emph{to some degree}. In fact, the image enhancement pipeline might recover real details in the photographed scene that would not be visible on an image obtained from raw sensor data. These details are certainly not recovered through a ``humanely mediated'' contact with the scene, as would be the case if someone was digitally painting over the photograph. But they also stretch the definition of what Walton calls a ``mechanical connection'' with reality. Indeed, details are added because the enhancement algorithm has learned the approximate probability distribution of a very large number of other photographs contained in its training data. Added details might fall within the correct distribution, yet differ from the ground truth in significant ways (see fig. \ref{fig:SR} for an extreme example). The complete process -- from the training to the deployment of a DL algorithm on a particular photograph -- could be described as mechanistic, but the ``contact'' between the final output and the depicted scene is mediated, in some way, by the ``contact'' between millions of training samples and the scenes they depict.

Walton also highlights that manipulated photographs may \emph{appear} to be transparent in respects in which they are not (\cite[][pp. 44]{waltonTransparentPicturesNature1984}). This is all the more true for DLSAM given the degree of photorealism they can achieve. A convincing face-swapping deepfake may be indistinguishable from a genuine video, hence their potential misuse for slandering, identity theft, and disinformation. Does one really ``see'' Tom Cruise when watching a deepfake video of an actor whose face has been replaced with that of Tom Cruise (fig. \ref{fig:deepfakes})? Not quite -- but the answer is complicated. The output does faithfully reflect the features of Tom Cruise's face, based on the summary statistics captured by the autoencoder architecture during training on the basis of genuine photographs of that face. Furthermore, this process is not mediated by the intentional attitudes of the deepfake's creator, in the way in which making a painting or a 3D model of Tom Cruise would be. A totally synthetic GAN-generated face is a more extreme case; yet it, too, is mechanically generated from the learned probability distribution of real photographs of human faces. Neither of these examples appear to be completely opaque in Walton's sense, although they can certainly be very deceiving in giving the illusion of complete transparency.

These remarks highlight the difficulty in drawing a clear boundary between traditional taxonomic categories when it comes to DLSAM. While traditional media do include a few edge cases, these are exceptions rather than the rule. The remarkable capacity of DL models to capture statistically meaningful properties of their training data and generate convincing samples that have similar properties challenges the divide between reality and synthesis, as well as more fine-grained distinctions between kinds of media synthesis. 

\subsection{Manifold learning and disentanglement}

The way in which deep generative models learn from data has deeper implications for the Continuity Question. According to the \emph{manifold hypothesis}, real-world high-dimensional data tend to be concentrated in the vicinity of low-dimensional manifolds embedded in a high-dimensional space (\cite{tenenbaumGlobalGeometricFramework2000,carlssonTopologyData2009,feffermanTestingManifoldHypothesis2016}). Mathematically, a manifold is a topological space that locally resembles Euclidean space; that is, any given point on the manifold has a neighborhood within which it appears to be Euclidean. A sphere is an example of a manifold in three-dimensional space: from any given point, it locally appears to be a two-dimensional plane, which is why it has taken humans so long to figure out that the earth is spherical rather than flat. 

In the context of research on deep learning, a manifold refers more loosely to a set of points that can be approximated reasonably well by considering only a small numbers of dimensions embedded in a high-dimensional space (\cite{Goodfellow-et-al-2016}). DL algorithms would have little chance of learning successfully from $n$-dimensional data if they had to fit a function with interesting variations across every dimension in $\mathbb{R}^{n}$. If the manifold hypothesis is correct, then DL algorithm can learn much more effectively by fitting low-dimensional nonlinear manifolds to sampled data points in high-dimensional spaces -- a process known as \emph{manifold learning}. 

There are theoretical and empirical reasons to believe that the manifold hypothesis is at least approximately correct when it comes to many kinds of data fed to DL algorithms, including audiovisual media. First, the probability distribution over real-world sounds and images is highly concentrated. Sounds and images are intricately structured in real life, and span a low-dimensional region of the high-dimensional space in which they are embedded. As we have seen, a 512x512 image can be represented as vector in a space with 786,432 dimensions -- one for each RGB channel for each pixel. Suppose that we generate such an image by choosing random color pixel values; the chance of obtaining anything that looks remotely different from uniform noise is absurdly small. This is because the probability distribution of real-world 512x512 images (i.e., images that mean something to us) is concentrated in a small region of $\mathbb{R}^{786432}$, on a low-dimensional manifold.\footnote{This toy example ignores the fact that natural images may lie on a union of disjoint manifolds rather than one globally
connected manifold. For example, the manifold of images of human faces may not be connected to the manifold of images of tropical beaches.} The same applies to sound: a randomly generated audio signal has a much greater chance of sounding like pure noise than anything meaningful. 

We can also intuitively think of transforming audiovisual media within a constrained region of their input space. For example, variations across real-world images can be boiled down to changes along a constrained set of parameters such as brightness, contrast, orientation, color, etc. These transformations trace out a manifold in the space of possible images whose dimensionality is much lower than the number of pixels, which lends further credence to the manifold hypothesis. Consequently, we can expect an efficient DL algorithm trained on images to represent visual data in terms of coordinates on the low-dimensional manifold, rather than in terms of coordinates in $\mathbb{R}^{n}$ (where $n$ is three times the number of pixels for a color image).

Deep generative models used for audiovisual media synthesis are good examples of manifold learning algorithms. For example, the success of GANs in generating images that share statistically relevant properties with training samples (e.g., photorealistic images of human faces) can be explained by the fact that they effectively discover the distribution of the dataset on a low-dimensional manifold embedded in the high-dimensional input space. This is apparent when interpolating between two points in the latent space of a GAN, namely traversing the space from one point to the other along the learned manifold: if the image corresponding to each point is visualized as a video frame, the resulting video shows a smooth -- spatially and semantically coherent -- transformation from one output to another (e.g., from one human face to another).\footnote{See \url{https://youtu.be/6E1_dgYlifc} for an example of video interpolation in the latent space of StyleGAN trained on photographs of human faces.}

Similarly, the autoencoder architecture of traditional ``deepfakes'' used for video face-swapping is a manifold learning algorithm. Over the course of training, it extracts latent features of faces by modeling the natural distribution of these features along a low-dimensional manifold (\cite{bengioRepresentationLearningReview2013}). The autoencoder's success in mapping one human face to another, in a way that is congruent with head position and expression, is explained by the decoder learning a mapping from the low-dimensional latent space to a manifold embedded in high-dimensional space (e.g., pixel space for images) (\cite{shaoRiemannianGeometryDeep2017}).

Thus, deep generative models used for audiovisual media synthesis effectively learn the distribution of data along nonlinear manifolds embedded in high-dimensional input space. This highlights a crucial difference between DLSAM and traditional synthetic media. Traditional media manipulation and synthesis is mostly \emph{ad hoc}: it consists in transforming or creating sounds, images, and videos in a specific way, with a specific result in mind. Many of the steps involved in this process are discrete manipulations, such as removing a portion of a photograph in an image editing software, or adding a laughing track to a video. These manipulations are tailored to a particular desired output. By contrast, DLSAM are sampled from a continuous latent space that has not been shaped by the desiderata of a single specific output, but by manifold learning. Accordingly, synthetic features of DLSAM do not originate in discrete manipulations, but from a mapping between two continuous spaces -- a low-dimensional manifold and a high-dimensional input space. This means that in principle, synthetic features of DLSAM can be altered as continuous variables. This is also why one can smoothly interpolate between two images within the latent space of a generative model, whereas it is impossible to go from an image to another through a continuous transformation with an image editing software.

Beyond manifold learning, recent generative models have been specifically trained to learn \emph{disentangled representations} (\cite{shenInterpretingLatentSpace2020,collinsEditingStyleUncovering2020,harkonenGANSpaceDiscoveringInterpretable2020,wuStyleSpaceAnalysisDisentangled2021}). As a general rule, the dimensions of a model's latent space do not match neatly onto interpretable features of the data. For example, shifting the vector corresponding to a GAN-generated image along a particular dimension of the generator's latent space need not result in a specific visual change that clearly corresponds to some particular property of the depiction, such as a change in the orientation of the subject's face for a model trained on human faces. Instead, it might result in a more radical visual change in the output, where few features of the original output are preserved. Disentanglement loosely refers to a specific form of manifold learning in which each latent dimension controls a single visual attribute that is interpretable. Intuitively, disentangled dimensions capture distinct generative factors -- interpretable factors that characterize every sample from the training data, such as the size, color, or orientation of a depicted object (\cite{bengioRepresentationLearningReview2013,higginsDefinitionDisentangledRepresentations2018}).

The advent of disentangled generative models has profound implications for the production of DLSAM. Disentanglement opens up new possibilities for manipulating any human-interpretable attribute within latent space. For example, one could generate a photorealistic image of a car, then manipulate specific attributes such as color, size, type, orientation, background, etc. Each of these disentangled parameters can be manipulated as continuous variables within a disentangled latent space, such that one can smoothly interpolate between two outputs along a single factor -- for example, going from an image of a red car seen from the left-hand side to an image of an identical red car seen from the right-hand side, with a smooth rotation the vehicle, keeping all other attributes fixed. Disentangled representations can even be continuously manipulated with an easy user interface, such as sliders corresponding to each factor (\cite{harkonenGANSpaceDiscoveringInterpretable2020,abdalStyleFlowAttributeconditionedExploration2020}).\footnote{See \url{https://www.nytimes.com/interactive/2020/11/21/science/artificial-intelligence-fake-people-faces.html} for interactive examples of disentangled GAN interpolation with images of human faces.}

Combined with aforementioned methods to ``invert'' a real image within the latent space of a generative model, disentanglement is becoming a novel and powerful way to manipulate pre-existing visual media, including photographs, with impressive precision. Thus, the manipulation of simple and complex visual features in fig. \ref{fig:russell} is made not only possible but trivial with a well-trained disentangled GAN. With domain-general generative models such as BigGAN (\cite{brockLargeScaleGAN2019}), the combination of inversion and disentanglement will soon allow nontechnical users to modify virtually any photograph along meaningful continuous dimensions. Multimodal Transformer models trained on text-image pairs like CLIP make this process even easier by allowing users to simply describe in natural language the change they would like to see effected in the output, while specifying the magnitude of the desired manipulation (fig. \ref{fig:russell}, \cite{patashnikStyleCLIPTextDrivenManipulation2021}).

Beyond static visual manipulation, ``steering'' the latent space of a disentangled generative model has the potential to allow any image to be animated in semantically coherent ways (\cite{jahanianSteerabilityGenerativeAdversarial2020}). There is, in principle, no difference between dynamically steering the latent space of a ``static'' generative model, and generating a photorealistic video. For example, one could invert the photograph of a real human face into latent space, then animate it by steering the space along disentangled dimensions -- moving the mouth, eyes, and entire head in a natural way. Thus, the task of video synthesis can now be reduced to discovering a trajectory in the latent space of a fixed image generator, in which content and motion are properly disentangled (\cite{tianGoodImageGenerator2021}). State-of-the-art methods to remove texture inconsistencies during interpolation demonstrate that latent space steering can be virtually indistinguishable from real videos (\cite{karrasAliasFreeGenerativeAdversarial2021}). In the near future, it is likely that any image or photograph can be seamlessly animated through this process, with congruent stylistic attributes -- from photorealistic to artistic and cartoonish styles. 

The capacity to steer generative models along interpretable dimensions in real time also paves the way for a new kind of synthetic medium: controllable videos that we can interact with in the same robust way that we interact with video games (\cite{menapacePlayableVideoGeneration2021,kimDriveGANControllableHighQuality2021}). For example, one can use a keyboard to move a tennis player forward, backward, leftward, and rightward on a field in a synthetic video. One can imagine that more robust models will soon enable similar real-time control over generated videos for complex variations and movements. Furthermore, one could invert a real photograph within the latent space of a generative model, then animate elements of the photograph in real-time in a synthetic video output. This kind of synthetic media pipeline has no equivalent with more traditional methods. The interactive nature of the resulting DLSAM is only matched by traditional video games, but these still fall short of the photorealism achieved by deep generative models, lack the versatility of what can be generated from the latent space of a domain-general model, and cannot be generated from pre-existing media such as photographs without significant processing and manual labor (e.g. with photogrammetry).

Manifold learning and disentanglement play an important role in setting some DLSAM apart from traditional synthetic audiovisual media. They further blur the line between the archival and the synthetic domain, since well-trained generative models capture the dense statistical distribution of their training data, and can seamlessly produce new sample or reconstitute existing media from that learned distribution. Disentanglement allows fine-grained control over the output of such models along specific interpretable dimensions, creating unforeseen possibility for media manipulation and real-time synthesis, with many more degrees of freedom than what was possible from previous techniques. 

\section{Conclusion}
\label{conclu}

The progress of deep learning algorithms is changing the way in which audiovisual media can be produced and manipulated. This change is more significant than the shift from analog to digital production tools. DL-based synthetic audiovisual media, including original deepfakes, require a lot less time, artistic skills, and -- increasingly -- technical expertise and computational resources to produce. They also greatly surpass traditional techniques in many domains, particularly for the creation and manipulation of realistic sounds, images, and videos. 

Beyond these incremental improvements, however, DLSAM represent a genuine departure from previous approaches that opens up new avenues for media synthesis. Manifold learning allows deep generative models to learn the probability distribution of millions of samples in a given domain, and generate new samples that fall within the same distribution. Disentanglement allows them to navigate the learned distribution along human-interpretable generative factors, and thus to manipulate and generate high-quality media with fine-grained control over their discernible features. Unlike traditional methods, the generative factors that drive the production of DSLAM exist on a continuum as dimensions of the model's latent space, such that any feature of the output can in principle be altered as a continuous variable. 

These innovations blur the boundary between familiar categories of audiovisual media, particularly between archival and synthetic media; but they also pave the way for entirely novel forms of audiovisual media, such as controllable images and videos that can be navigated in real-time like video games, or multimodal generative artworks (e.g., images and text jointly sampled from the latent space of a multimodal model). This warrants treating DL-based approaches as a genuine paradigm shift in media synthesis. While this shift does have concerning ethical implications for the potentially harmful uses of DLSAM, it also opens up exciting possibilities for their beneficial use in art and entertainment.\footnote{I am grateful to Paul Egré and Benjamin Icard for helpful feedback on an early draft of this paper, as well as comments from two anonymous referees.}

\bibliographystyle{unsrtnat}
% \bibliography{bibliography}

\end{document}